\documentclass{article}
\usepackage{booktabs}
\usepackage[margin=1in]{geometry}
\usepackage{adjustbox}
\usepackage{multirow}
\usepackage{spconf,amsmath,graphicx,hyperref}
\usepackage{amssymb}
\usepackage{float} 
\usepackage{color}

\title{RareBoost3D: A synthetic Lidar Dataset with Enhanced Rare Classes}
\name{Shutong Lin \qquad Zhengkang Xiang \qquad Jianzhong Qi \qquad Kourosh Khoshelham\thanks{The dataset and code will be released after paper acceptance.}}
\address{
	The University of Melbourne\\
     Australia}

\begin{document}

\maketitle


\begin{abstract}
Real-world point cloud datasets have made significant contributions to the development of LiDAR-based perception technologies, such as object segmentation for autonomous driving. However, due to the limited number of instances in some rare classes, the long-tail problem remains a major challenge in existing datasets. To address this issue, we introduce a novel, synthetic point cloud dataset named \emph{RareBoost3D}, which complements existing real-world datasets by providing significantly more instances for object classes that are rare in real-world datasets. To effectively leverage both synthetic and real-world data, we further propose a cross-domain semantic alignment method named \emph{CSC} loss that aligns feature representations of the same class across different domains. Experimental results demonstrate that this alignment significantly enhances the performance of LiDAR point cloud segmentation models over real-world data.
\end{abstract}

\section{Introduction}
\label{sec:intro}

\begin{figure}[t]
  \centering
  \includegraphics[width=0.9\linewidth]{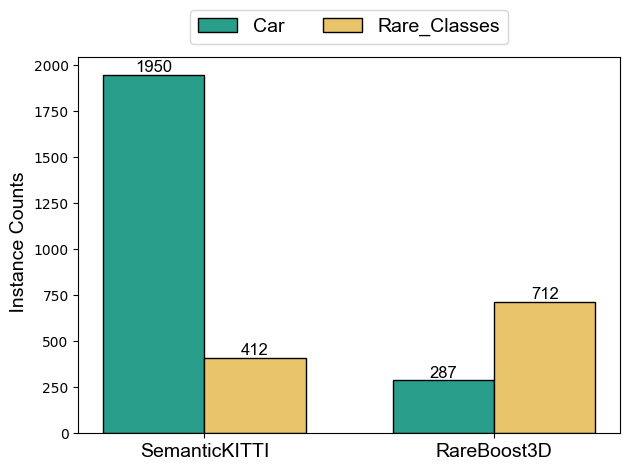}
  \caption{\footnotesize The numbers of instances for non-rare (\textit{car}) and rare  (\textit{person}, \textit{bicycle}, \textit{motorcycle}, \textit{rider}, and \textit{truck}) classes in SemanticKITTI~\cite{behley2019semantickitti} and our RareBoost3D. SemanticKITTI exhibits significant class imbalance, with the number of \textit{car} instances being roughly 5 times greater than the combined number of instances of the rare classes. In contrast, RareBoost3D has much more instances for the rare classes.}
  \label{fig:stat}
\end{figure}

Semantic segmentation of point clouds plays a crucial role in interpreting 3D environments, especially in applications such as autonomous driving, where LiDAR sensors capture rich geometric information. However, due to the sparse and unstructured nature of point clouds, obtaining high-quality annotations for real-world LiDAR sequences is both costly and time-consuming~\cite{behley2019semantickitti, xiao20233d}. Existing real-world point cloud datasets are often limited in terms of size, diversity, and coverage of objects in different categories (``\emph{classes}'')~\cite{xiang2024synthetic}. Figure~\ref{fig:stat} illustrates the class imbalance issue of a popular real-world dataset SemanticKITTI~\cite{behley2019semantickitti}.

To alleviate the issues posed by limited and imbalanced real-world datasets, data augmentation~\cite{srivastava2014dropout} is commonly adopted to enhance the diversity of training samples. However, most traditional augmentation methods are applied either globally to the entire point cloud scene or locally to individual objects~\cite{cheng2025object}. These methods primarily operate on the geometric level (e.g., rotating an object),  without introducing novel samples into the dataset. 

Inspired by the success of joint training across multiple datasets~\cite{wu2024towards} and transfer learning from synthetic to real domains~\cite{xiao2022transfer}, we introduce \emph{RareBoost3D}, a large-scale synthetic LiDAR sequential dataset generated using the open-source CARLA~\cite{dosovitskiy2017carla} simulator. This dataset provides point-wise semantic annotations in diverse urban and rural scenes, with more instances of objects in classes that are rare  (``\emph{rare class}'' hereafter) in real-world datasets -- see Figure~\ref{fig:stat} which plots the class distribution of RareBoost3D for a subset of the classes. It serves as an effective data augmentation resource to alleviate the class imbalance issues of real-world datasets.

While synthetic data can help reduce labeling costs, there could be a significant domain gap between synthetic and real-world LiDAR data~\cite{xiang2025sgldm}, due to difficulties in accurately simulating real-world textures and sensor noise. Synthetic point clouds tend to be geometrically smoother, which can lead to performance drop when a segmentation model is trained on synthetic data and applied to real-world scenarios. Several existing methods, e.g., SynLiDAR~\cite{xiao2022transfer} and ePointDA~\cite{zhao2021epointda}, attempt to mitigate this domain gap by simulating the appearance and sensor noise of real-world point clouds through adversarial learning. 

Unlike existing solutions, which often require complex adversarial training, we use contrastive learning~\cite{khosla2020supervised} to align features from synthetic and real-world domains in a shared semantic embedding space. We propose a prototype-based \emph{Cross-domain Semantic Consistency} (CSC) loss to guide the alignment process. Our experiments demonstrate that this alignment improves the generalization of segmentation models to real-world point clouds. 

Overall, this paper makes the following contributions: (1) We present RareBoost3D, a large-scale synthetic LiDAR dataset that offers diverse urban and rural scenes, enriched with more instances of rare classes. (2) We propose a prototype-based loss named CSC loss that leverages contrastive learning to align semantic features from synthetic and real domains in a shared embedding space.(3) Experimental results show that our method outperforms traditional augmentation methods, achieving around 2\% to 3\% improvement in segmentation performance. Through further analysis, we find that adjusting the distribution of rare classes in synthetic data can significantly boost their segmentation performance.


\section{Related Work}
\label{sec:related_work}




Constructing large-scale real-world point cloud datasets is costly and labor intensive. Inspired by the success of 2D synthetic datasets (e.g., SYNTHIA~\cite{ros2016synthia}) for pixel-level semantic segmentation tasks~\cite{park2022dat}, researchers have introduced synthetic point cloud datasets such as CarlaScenes~\cite{kloukiniotis2022carlascenes}. However, the scale of these datasets remains relatively small. A more recent effort, SynLiDAR~\cite{xiao2022transfer}, introduced a larger synthetic dataset, though it does not specifically address the challenge of rare classes. Another work similar to ours is SynthmanticLiDAR~\cite{montalvo2024synthmanticlidar}, which also generates synthetic point clouds using the CARLA simulator. Its goal is to replicate the class distribution of real-world datasets like SemanticKITTI, which suffers from the imbalance class issue. In contrast, our dataset not only offers large scale but also exploits the high controllability of the virtual environment to increase the frequency of objects in rare classes. This helps mitigate the class imbalance problem commonly observed in existing point cloud datasets~\cite{behley2019semantickitti, caesar2020nuscenes}.

\section{The RareBoost3D Dataset}
\label{sec:RareBoost3D}
Our RareBoost3D data is created using CARLA (v0.9.15)~\cite{dosovitskiy2017carla}, an open-source simulator built on the Unreal Engine. CARLA simulates realistic driving scenarios and enables data collection through virtual autonomous driving sensors (e.g., LiDAR, radar, and depth cameras) to support related research.

We use a simulated LiDAR sensor to collect point cloud sequences from eight different maps. These maps cover diverse outdoor environments, including rural areas (Map 7), small towns (Maps 1, 2, and 4), and large cities (Maps 3, 5, 6, and 10). The simulated sensor closely resembles  Velodyne HDL-64E~\cite{behley2019semantickitti}, which generates point clouds by emitting and receiving laser beams from 64 distinct vertical channels. Each scan returns the 3D coordinates of points along with their corresponding semantic labels assigned by the simulator. Since the virtual environments in CARLA cannot accurately simulate the real-world reflectance intensity~\cite{xiao2022transfer}, only point coordinates are used as input features in this paper. The resulting dataset includes a total of 29 semantic labels, where label 0 is reserved for ``unlabeled points''. The complete RareBoost3D dataset consists of eight LiDAR point cloud sequences, each corresponding to a different map and annotated at the point level. Each sequence contains 60,000 scans, with each scan comprising approximately 125,000 to 138,000 points.



To address the long-tail problem of rare classes in real-world datasets, we constructed our dataset by adjusting the proportions of object instances that are rare in real datasets to create a more balanced  class distribution. The resulting dataset includes overall augmentations   across multiple rare classes (including \textit{person}, \textit{bicycle}, \textit{motorcycle}, \textit{rider}, and \textit{truck}), as well as augmentations for targeted individual classes. For example, certain subsets contain more than 1,000 pedestrian instances or up to 367 trucks. This design enables users to selectively strengthen the representation learning for instances of rare class in existing real-world datasets. In this paper, we focus on evaluating the overall improvements in all rare classes. Our experiments are conducted on subsets in which all rare classes have been proportionally enhanced.

\section{Cross-domain Semantic Consistency}
\label{sec:cdsc}

To better learn domain-invariant representations from both real-world and synthetic point cloud datasets, we adapt a cross-domain feature alignment method from PointDR~\cite{xiao20233d}. This method employs contrastive learning~\cite{khosla2020supervised, li2025openworld, li2024contrastive} to pull together representations of instances from the same class while pushing apart those from different classes. We construct class-wise feature prototypes for both real and synthetic datasets, denoted as \( f_{\text{real}} \in \mathbb{R}^{D \times C} \) and \( f_{\text{syn}} \in \mathbb{R}^{D \times C} \), where \( D \) is the feature dimensionality and \( C \) is the number of semantic classes. These prototypes are stored in two separate memory banks, \( B_{\text{real}} \) and \( B_{\text{syn}} \), which serve as keys during contrastive learning.

During training of a representation learning and semantic segmentation model, each point feature embedding \( q_i \) from the combined dataset (including both real and synthetic samples) is treated as a query, while the class prototypes stored in the two memory banks act as keys. The contrastive loss is computed by measuring the similarity between the query and prototypes of the same class, and the dissimilarity with prototypes of different classes from both domains. Specifically, the key sharing the same semantic class with the query is referred to as the positive key \( B^+ \), and the remaining class prototypes are treated as negative keys, denoted as \( B^j \). We use the subscripts ``real'' and ``syn'' to indicate the domain associated with each memory bank. The contrastive losses for both domains are defined as follows:

{\small
\begin{equation}
\mathcal{L}_{rc} = -\frac{1}{N} \sum_{i=1}^N \log \frac{\exp\left(q_i^\top \mathcal{B}^{+}_\text{real} / \tau \right)}{\sum_{j=1}^C \exp\left(q_i^\top \mathcal{B}^{j}_\text{real} / \tau \right)} \\
\label{eq:rc_loss}
\end{equation}
}

{\small
\begin{equation}
\mathcal{L}_{sc} = -\frac{1}{N} \sum_{i=1}^N \log \frac{\exp\left(q_i^\top \mathcal{B}^{+}_\text{syn} / \tau \right)}{\sum_{j=1}^C \exp\left(q_i^\top \mathcal{B}^{j}_\text{syn} / \tau \right)}
\label{eq:sc_loss}
\end{equation}
}
where $N$ is the number of training samples, and $\tau$ is a temperature coefficient. To jointly optimize for accurate semantic segmentation and cross-domain feature consistency, we combine the segmentation loss \( \mathcal{L}_{seg} \) with the two contrastive losses proposed above. The overall training objective is:

{\small
\begin{equation}
\mathcal{L}_{\text{csc}} = \mathcal{L}_{\text{seg}} +\mathcal{L}_{rc} + \mathcal{L}_{sc}
\end{equation}
}

\begin{table*}[t]
\centering
\caption{\footnotesize Overall performance results on the SemanticKITTI validation set. Combining RareBoost3D with SemanticKITTI leads to improved segmentation accuracy. The table shows a comparison of per-class IoU and mIoU scores for MinkUNet and PTV3. The bold and underlined values indicate the best and second-best results, respectively. Results denoted by `*' are adopted from~\cite{xiao2022transfer}. The \textbf{Ours} method here uses the same subset as in Table~\ref{tab:synthetic_settings}(Setting 1), with the CSC loss defined in Table~\ref{tab:ablation_study}.}
\label{tab:baseline_methods}
\setlength{\tabcolsep}{2pt}
\begin{adjustbox}{max width=\textwidth}
\begin{tabular}{llcccccccccccccccc|c}
\toprule
\textbf{Backbone} & \textbf{Method} & \textbf{car} & \textbf{road} & \textbf{build.} & \textbf{person} & \textbf{bi.cle} & \textbf{mt.cle} & \textbf{rider} & \textbf{truck} & \textbf{sidew.} & \textbf{fence} & \textbf{veget.} & \textbf{terra.} & \textbf{pole} & \textbf{traffic-sign} & \textbf{oth-g.} & \textbf{oth-v.} & \textbf{Val mIoU} \\
\midrule
\multirow{6}{*}{MinkUNet}
& Baseline & 94.4 & 93.0 & 91.1 & 57.0 & 2.6 & 43.7 & 79.7 & 64.2 & 78.3 & 53.1 & \underline{88.7} & 72.7 & 61.6 & 45.8 & 42.4 & 34.0 & 62.6 \\
& Dropout & 94.5 & 93.1 & 91.2 & 54.7 & 3.7 & 45.8 & 79.4 & 71.4 & 79.0 & 53.9 & \textbf{89.3} & \textbf{74.5} & 61.4 & 43.8 & \textbf{45.2} & 34.1 & 63.4 \\
& Jittering & 94.5 & 92.9 & \underline{91.3} & 59.2 & 3.0 & 51.2 & \underline{81.1} & \underline{82.4} & 78.5 & 53.5 & 88.7 & 72.7 & 61.3 & 44.1 & 42.2 & 36.8 & 64.6 \\
& PointAug* & \underline{95.9} & \textbf{93.8} & 89.8 & 67.0 & \underline{29.2} & \textbf{70.0} & - & 76.3 & \textbf{81.2} & \underline{58.4} & 87.5 & 72.7 & \underline{62.4} & \textbf{50.5} & 4.6 & \underline{50.0} & \underline{66.0} \\
& SynLiDAR* & \textbf{95.9} & 92.3 & 89.8 & \textbf{71.4} & \textbf{33.0} & \underline{62.8} & - & 78.9 & \underline{79.9} & \textbf{59.5} & 86.3 & \underline{72.8} & \textbf{63.6} & \underline{48.9} & 0.1 & \textbf{50.2} & 65.7 \\
& \textbf{Ours} & 95.4 & \underline{93.3} & \textbf{91.7} & \underline{69.3} & 11.4 & 56.0 & \textbf{85.1} & \textbf{82.7} & 79.0 & 55.2 & 88.6 & 71.9 & 61.9 & 47.2 & \underline{45.2} & 48.6 & \textbf{67.7} \\

\midrule
\multirow{4}{*}{PTV3} 
& Baseline & 95.1 & \underline{93.7} & \underline{91.7} & 67.0 & \underline{9.6} & \underline{61.6} & \underline{82.3} & 78.5 & 79.4 & \underline{56.6} & \underline{88.5} & \underline{71.6} & 62.3 & 48.8 & 40.1 & 48.2 & 67.2 \\
& Dropout & \underline{95.2} & 93.7 & 91.1 & \underline{68.1} & 7.6 & 61.1 & 80.6 & \underline{81.8} & 79.4 & 53.0 & 88.4 & 71.1 & \underline{62.9} & 50.0 & \textbf{42.3} & \underline{50.0} & \underline{67.3} \\
& Jittering & 95.0 & 93.6 & 91.2 & 67.9 & 7.9 & 61.1 & 80.4 & 77.5 & \underline{79.5} & 53.8 & \textbf{88.7} & \textbf{72.1} & 62.5 & \underline{50.5} & 40.3 & 49.0 & 66.9 \\
& \textbf{Ours} & \textbf{95.7} & \textbf{94.2} & \textbf{92.1} & \textbf{71.2} & \textbf{14.0} & \textbf{66.2} & \textbf{83.0} & \textbf{86.1} & \textbf{80.7} & \textbf{57.2} & 88.4 & 70.8 & \textbf{63.3} & \textbf{51.7} & \underline{42.1} & \textbf{53.4} & \textbf{69.4} \\
\bottomrule
\end{tabular}
\end{adjustbox}
\end{table*}

\begin{table}[t]
\centering
\caption{\footnotesize Performance results when increasing the number of instances in rare classes. The number of instances in setting 1 and 2 represent the count of each rare class in the RareBoost3D subsets, which should be added to the corresponding classes in the baseline to achieve the performance reported for each setting.}


\label{tab:synthetic_settings}
\setlength{\tabcolsep}{3pt}
{\small
\begin{adjustbox}{max width=\textwidth}
\begin{tabular}{lcccccc|c}
\toprule
\textbf{Setup} & \textbf{car} & \textbf{person} & \textbf{bi.cle} & \textbf{mt.cle} & \textbf{rider} & \textbf{truck} & \textbf{mIoU} \\
\midrule
Baseline & \underline{95.1} & 67.0 & 9.6 & 61.6 & 82.3 & 78.5 & 65.7\\
Instances & 1950 & 139 & 128 & 57 & 60 & 28 & - \\
\midrule
Setting 1 & \textbf{95.2} &  \underline{70.0} & \underline{15.3} & \underline{68.3} & \underline{83.1} & \textbf{85.4} & \underline{69.6}\\
Instances & 287 & 198 & 64 & 117 & 180 & 153 & - \\
\midrule
Setting 2 & 94.9  & \textbf{71.0} & \textbf{19.0} & \textbf{69.5} & \textbf{84.6} & \underline{81.1} & \textbf{70.0}\\
Instances & 436 & 222 & 164 & 178 & 342 & 40 & - \\

\bottomrule
\end{tabular}
\end{adjustbox}
} 

\end{table}

\section{Experiments}
\label{sec:experiments}






\subsection{Experimental Setup}

\noindent\textbf{Dataset.} All experiments are conducted with RareBoost3D and the \textbf{SemanticKITTI} dataset~\cite{behley2019semantickitti}, a large-scale real-world LiDAR point cloud benchmark for outdoor semantic segmentation. It contains 43,552 densely annotated scans with point-wise labels across 25 semantic classes. We follow the official split and use 19,130 scans from sequences 00 to 07 and 09 to 10 for training, 4,071 scans from sequence 08 for validation, and 20,351 scans from the remaining sequences for testing.

\noindent\textbf{Backbone network.} To evaluate the generalizability of our dataset across different backbone architectures, we conduct experiments using two networks of significantly different designs: (1) MinkUNet~\cite{choy20194d} is based on sparse convolution and provides high computational efficiency; and (2) Point Transformer V3 (\textbf{PTV3})~\cite{wu2024point} is based on self-attention and demonstrates strong capability in modeling global dependencies.

\noindent\textbf{Training details.} All experiments are run on four Nvidia A100 GPUs (80 GB memory each). We train all models for 50 epochs with a batch size of 16. We use the AdamW optimizer with an initial learning rate of 2e-4 and a weight decay of 0.005. The learning rate is scheduled using the OneCycleLR policy and linearly increases to 2e-3 during the first few training steps, followed by a cosine annealing schedule.

\noindent\textbf{Label mapping.} The semantic label sets provided by CARLA and SemanticKITTI  are not directly aligned. We unify them by mapping both to a common set of semantic classes. For SemanticKITTI, we merge the original labels \{\textit{bicyclist}, \textit{motorcyclist}\} into the \textit{rider} class, \{\textit{parking}, \textit{other-ground}\} into \textit{other-ground}, and \{\textit{vegetation}, \textit{trunk}\} into \textit{vegetation}. For RareBoost3D, class that are either rarely collected in the real-world environment (e.g., \textit{sky}, \textit{water}) or cannot be aligned with the common label set (e.g., \textit{trains}) are ignored during training. After label alignment, we retain a total of 16 unified semantic classes: \textit{car}, \textit{road}, \textit{building} (build.), \textit{person}, \textit{bicycle} (bi.cle), \textit{motorcycle} (mt.cle), \textit{rider}, \textit{truck}, \textit{sidewalk} (sidew.), \textit{fence}, \textit{vegetation} (veget.), \textit{terrain} (terra.), \textit{pole}, \textit{traffic-sign}, \textit{other-ground} (oth-g.), and \textit{other-vehicle} (oth-v.).

\noindent\textbf{Evaluation metrics.} We report the Intersection over Union (IoU) and mean Intersection over Union (mIoU). IoU measures the overlap between the predicted and ground truth segmentation region for each class, while the mIoU represents the average IoU across all classes, providing an overall assessment of the model's segmentation accuracy.

\subsection{Results for Overall Data Augmentation}
We begin by evaluating the effectiveness of using RareBoost3D to augment real-world point cloud data compared to other data augmentation methods. As shown in Table~\ref{tab:baseline_methods}, the comparison includes three baseline methods: (1) No data augmentation, which trains the backbone models with the raw SemanticKITTI dataset;  (2) Random dropout~\cite{srivastava2014dropout}, which randomly removes 20 percentage of points from each scan to simulate missing data; (3) Random jittering, which adds Gaussian noise to each point’s coordinates to simulate sensor noise, and two advanced augmentation methods: (4) PointAug~\cite{li2020pointaugment} and (5) SynLiDAR~\cite{xiao2022transfer}. Note that the results for the last two methods are obtained from SynLiDAR~\cite{xiao2022transfer} which uses the MinkUNet backbone.

Table~\ref{tab:baseline_methods} shows that using RareBoost3D to augment the model training set (i.e., \textbf{Ours}) improves model performance. For example, the two backbone networks achieve 3\% to 10\% improvements on rare classes such as \textit{person}, \textit{bicycle} and \textit{motorcycle} comparing with three baseline methods. Compared with SynLiDAR and PointAug, using RareBoost3D also achieves better performance on rare classes such as \textit{person} and \textit{truck}. This suggests that our synthesized samples for these classes better match the real-world data distribution. For other rare classes, such as \textit{bicycle} and \textit{motorcycle}, using RareBoost3D  still lags behind PointAug and SynLiDAR. This may be attributed to the limited shape diversity of bicycles and motorcycles in our dataset.

In addition, we observe a slight performance drop in certain background classes, such as \textit{terrain} and \textit{vegetation}. This may be attributed to the fact that background classes in real-world environments often contain subtle texture details, which are typically oversimplified in synthetic data. 



\subsection{Impact of Proportion of Instances in Rare Classes}

To further evaluate the effectiveness of using RareBoost3D to enhance segmentation performance in rare classes, we vary the proportions of instances in rare classes. We compare with model training solely on SemanticKITTI. As shown in Figure~\ref{fig:stat}, SemanticKITTI contains 1,950 \textit{car} instances (a frequent class) and only 412 instances from rare categories (including \textit{person, bicycle, motorcycle, rider, and truck}). In \textbf{Setting~1}, we augment SemanticKITTI with a RareBoost3D subset enriched with instances in those rare classes, increasing their total number to 1,124. In \textbf{Setting~2}, we replace this subset with another that contains an even higher proportion of instances in those rare classes, i.e., a total to 1,358.

As shown in Table~\ref{tab:synthetic_settings}, the results of {Setting~1} demonstrate that RareBoost3D helps improve the segmentation accuracy  in all five rare classes. In {Setting~2}, where the number of rare-class instances was further increased, we observe even greater performance gains for all rare classes, except for the \textit{truck} class. This discrepancy is mainly due to the fact that {Setting~1} contained approximately 181 truck instances, whereas {Setting~2} had only around 68. This reduced number of \textit{truck} instances in {Setting~2} likely limited the model's ability to learn this class effectively.

Meanwhile, both {Setting~1} and {Setting~2} also introduced a number of \textit{car} instances, with 287 and 436 respectively. However, the model's performance on this frequent class remained largely unchanged. This suggests that augmenting frequent classes offers limited benefit, and highlights the value of the RareBoost3D dataset in significantly improving the representation of rare categories. Overall, these results demonstrate the effectiveness of RareBoost3D in enhancing model performance on rare categories, particularly under conditions of severe class imbalance.

\begin{table}[t]
\centering
\caption{\normalsize Ablation study results for the CSC loss.}
\label{tab:ablation_study}

\normalsize
\begin{tabular}{ll|c}
\toprule
\textbf{Backbone} & \textbf{Method} & \textbf{Val mIoU} \\
\midrule
\multirow{2}{*}{MinkUNet}
& RareBoost3D + $\mathcal{L}_{\text{seg}}$ & 65.8 \\
& RareBoost3D + $\mathcal{L}_{\text{csc}}$ & \textbf{67.7}\\
\midrule
\multirow{2}{*}{PTV3} 
& RareBoost3D + $\mathcal{L}_{\text{seg}}$ & 68.9 \\
& RareBoost3D + $\mathcal{L}_{\text{csc}}$ & \textbf{69.4}\\
\bottomrule
\end{tabular}
\end{table}

\subsection{Ablation Study}
To assess the effectiveness of the proposed CSC loss, we conduct ablation studies under two training settings: (1) training with RareBoost3D using cross-entropy and Lovasz loss as the training objectives ($\mathcal{L}_{\text{seg}}$); and (2) incorporating the CSC loss ($\mathcal{L}_{\text{csc}}$) in addition to these objectives. 

As shown in Table~\ref{tab:ablation_study}, without CSC loss, the two backbone networks report mIoU of 65.8\% and 68.9\%, respectively. After adding CSC loss, the mIoU improves to 67.7\% and 69.4\%. These results demonstrate that learning domain-invariant features through CSC loss effectively mitigates the impact of domain discrepancies and enhances segmentation accuracy.


\section{Conclusion} 
\label{sec:conclusion}
We presented RareBoost3D, a large-scale, high-fidelity synthetic point cloud dataset with point-wise annotations, designed to complement real-world datasets by increasing the presence of objects in rare semantic classes. We proposed a contrastive learning loss named CSC loss that helps align feature representations of the same semantic class across real and synthetic domains. Experimental results show that the rare classes in RareBoost3D share a high similarity to those in a popular real-world dataset SemantiKITTI, and their IoU improves as the number of instances increases. Together with CSC loss, RareBoost3D effectively improves the overall performance of segmentation models by approximately 1\%, depending on the type of backbone network used.



\vfill\pagebreak

\bibliographystyle{IEEEbib}
\bibliography{strings,refs}

\begin{thebibliography}{10}

\bibitem{behley2019semantickitti}
Jens Behley, Martin Garbade, Andres Milioto, Jan Quenzel, Sven Behnke, Cyrill Stachniss, and Jurgen Gall,
\newblock ``{SemanticKITTI}: A dataset for semantic scene understanding of {LiDAR} sequences,''
\newblock in {\em ICCV}, 2019, pp. 9297--9307.

\bibitem{xiao20233d}
Aoran Xiao, Jiaxing Huang, Weihao Xuan, Ruijie Ren, Kangcheng Liu, Dayan Guan, Abdulmotaleb El~Saddik, Shijian Lu, and Eric~P Xing,
\newblock ``{3D} semantic segmentation in the wild: Learning generalized models for adverse-condition point clouds,''
\newblock in {\em CVPR}, 2023, pp. 9382--9392.

\bibitem{xiang2024synthetic}
Zhengkang Xiang, Zexian Huang, and Kourosh Khoshelham,
\newblock ``Synthetic {LiDAR} point cloud generation using deep generative models for improved driving scene object recognition,''
\newblock {\em Image and Vision Computing}, vol. 150, pp. 105207, 2024.

\bibitem{srivastava2014dropout}
Nitish Srivastava, Geoffrey Hinton, Alex Krizhevsky, Ilya Sutskever, and Ruslan Salakhutdinov,
\newblock ``Dropout: A simple way to prevent neural networks from overfitting,''
\newblock {\em Journal of Machine Learning Research}, vol. 15, no. 1, pp. 1929--1958, 2014.

\bibitem{cheng2025object}
Haoran Cheng, Junkai Xu, Liang Peng, Zheng Yang, Xiaofei He, and Boxi Wu,
\newblock ``Object-level data augmentation for visual {3D} object detection in autonomous driving,''
\newblock in {\em ICASSP}, 2025, pp. 1--5.

\bibitem{wu2024towards}
Xiaoyang Wu, Zhuotao Tian, Xin Wen, Bohao Peng, Xihui Liu, Kaicheng Yu, and Hengshuang Zhao,
\newblock ``Towards large-scale 3d representation learning with multi-dataset point prompt training,''
\newblock in {\em CVPR}, 2024, pp. 19551--19562.

\bibitem{xiao2022transfer}
Aoran Xiao, Jiaxing Huang, Dayan Guan, Fangneng Zhan, and Shijian Lu,
\newblock ``Transfer learning from synthetic to real {LiDAR} point cloud for semantic segmentation,''
\newblock in {\em AAAI}, 2022, pp. 2795--2803.

\bibitem{dosovitskiy2017carla}
Alexey Dosovitskiy, German Ros, Felipe Codevilla, Antonio Lopez, and Vladlen Koltun,
\newblock ``{CARLA}: An open urban driving simulator,''
\newblock in {\em Annual Conference on Robot Learning}, 2017, pp. 1--16.

\bibitem{xiang2025sgldm}
Zhengkang Xiang, Zizhao Li, Amir Khodabandeh, and Kourosh Khoshelham,
\newblock ``{SG-LDM}: Semantic-guided lidar generation via latent-aligned diffusion,''
\newblock in {\em ICCV}, 2025.

\bibitem{zhao2021epointda}
Sicheng Zhao, Yezhen Wang, Bo~Li, Bichen Wu, Yang Gao, Pengfei Xu, Trevor Darrell, and Kurt Keutzer,
\newblock ``{ePointDA}: An end-to-end simulation-to-real domain adaptation framework for lidar point cloud segmentation,''
\newblock in {\em AAAI}, 2021, pp. 3500--3509.

\bibitem{khosla2020supervised}
Prannay Khosla, Piotr Teterwak, Chen Wang, Aaron Sarna, Yonglong Tian, Phillip Isola, Aaron Maschinot, Ce~Liu, and Dilip Krishnan,
\newblock ``Supervised contrastive learning,''
\newblock {\em NeurIPS}, pp. 18661--18673, 2020.

\bibitem{ros2016synthia}
German Ros, Laura Sellart, Joanna Materzynska, David Vazquez, and Antonio~M Lopez,
\newblock ``The {SYNTHIA} dataset: A large collection of synthetic images for semantic segmentation of urban scenes,''
\newblock in {\em CVPR}, 2016, pp. 3234--3243.

\bibitem{park2022dat}
Jinyoung Park, Minseok Son, Sumin Lee, and Changick Kim,
\newblock ``Dat: Domain adaptive transformer for domain adaptive semantic segmentation,''
\newblock in {\em ICIP}, 2022, pp. 4183--4187.

\bibitem{kloukiniotis2022carlascenes}
Andreas Kloukiniotis, Andreas Papandreou, Christos Anagnostopoulos, Aris Lalos, Petros Kapsalas, Duong-Van Nguyen, and Konstantinos Moustakas,
\newblock ``{CarlaScenes}: A synthetic dataset for odometry in autonomous driving,''
\newblock in {\em CVPR}, 2022, pp. 4520--4528.

\bibitem{montalvo2024synthmanticlidar}
Javier Montalvo, Pablo Carballeira, and {\'A}lvaro Garc{\'\i}a-Mart{\'\i}n,
\newblock ``Synthmanticlidar: A synthetic dataset for semantic segmentation on {LiDAR} imaging,''
\newblock in {\em ICIP}, 2024, pp. 137--143.

\bibitem{caesar2020nuscenes}
Holger Caesar, Varun Bankiti, Alex~H Lang, Sourabh Vora, Venice~Erin Liong, Qiang Xu, Anush Krishnan, Yu~Pan, Giancarlo Baldan, and Oscar Beijbom,
\newblock ``{nuScenes}: A multimodal dataset for autonomous driving,''
\newblock in {\em CVPR}, 2020, pp. 11621--11631.

\bibitem{li2025openworld}
Zizhao Li, Zhengkang Xiang, Joseph West, and Kourosh Khoshelham,
\newblock ``From open vocabulary to open world: Teaching vision language models to detect novel objects,''
\newblock in {\em British Machine Vision Conference}, 2025.

\bibitem{li2024contrastive}
Zizhao Li, Kourosh Khoshelham, and Joseph West,
\newblock ``Contrastive class anchor learning for open set object recognition in driving scenes,''
\newblock {\em Transactions on Machine Learning Research}, 2024.

\bibitem{choy20194d}
Christopher Choy, JunYoung Gwak, and Silvio Savarese,
\newblock ``{4D} spatio-temporal convnets: Minkowski convolutional neural networks,''
\newblock in {\em CVPR}, 2019, pp. 3075--3084.

\bibitem{wu2024point}
Xiaoyang Wu, Li~Jiang, Peng-Shuai Wang, Zhijian Liu, Xihui Liu, Yu~Qiao, Wanli Ouyang, Tong He, and Hengshuang Zhao,
\newblock ``{Point Transformer V3}: Simpler faster stronger,''
\newblock in {\em CVPR}, 2024, pp. 4840--4851.

\bibitem{li2020pointaugment}
Ruihui Li, Xianzhi Li, Pheng-Ann Heng, and Chi-Wing Fu,
\newblock ``{PointAugment}: an auto-augmentation framework for point cloud classification,''
\newblock in {\em CVPR}, 2020, pp. 6378--6387.

\end{thebibliography}

\end{document}